\documentclass[conference]{IEEEtran}
\usepackage{graphicx} 
\usepackage{mathrsfs} 
\usepackage{picins}
\usepackage{cite}
\usepackage[small,bf,centerlast]{caption} 
\usepackage{hyperref}
\hypersetup{
pdfstartview=FitH,
pdfpagemode = None,
linkcolor = red,
citecolor = green,
colorlinks = true
}

\usepackage{makeidx}  

\begin{document}

\newcommand{\romantext}[1]{\mathrm{#1}}
\newcommand{\du}{\smash{\lower1.4ex \hbox{\char34}}\kern-.2ex}
\newcommand{\hu}{\kern-.2ex\hbox{\char92}}
\newcommand{\uv}[1]{\hu#1''}
\newcommand{\ud}{\mathrm{d}}
\newcommand{\valkoitem}[1]{\textsc{#1}}
\newcommand{\mut}{\romantext{mut}}
\newcommand{\cross}{\romantext{cross}}
\newcommand{\desc}{\romantext{desc}}
\newcommand{\atan}{\romantext{atan}}
\newcommand{\PSP}{\romantext{PSP}}
\newcommand{\IMP}{\romantext{IMP}}
\newcommand{\MP}{\romantext{MP}}
\newcommand{\e}{\mathrm{e}}
\newcommand{\E}{\mathrm{E}}
\newcommand{\cv}{c_v}
\newcommand{\cvtheta}{c_{v_{\theta}}}
\newcommand{\ms}{\,\romantext{ms}}
\newcommand{\interval}[2]{\left\langle#1, #2\right\rangle}
\newcommand{\openinterval}[2]{\left(#1, #2\right)}
\newcommand{\lopeninterval}[2]{\left(#1, #2\right\rangle}
\newcommand{\ropeninterval}[2]{\left\langle#1, #2\right)}
\newcommand{\isi}{\overline{isi}} 
\newcommand{\Edis}{\mathscr{E}}
\newcommand{\Gdis}{\mathscr{G}}
\newcommand{\Pdis}{\mathscr{P}}
\newcommand{\Udis}{\mathscr{U}}
\newcommand{\valkonote}[1]{[\valkoitem{\textcolor{blue}{\bf #1}}]}
\newcommand{\nmmnote}[1]{[\valkoitem{\textcolor{red}{\bf #1}}]}
\newcommand{\pleasecheck}[1]{[\valkoitem{\textcolor{green}{\bf #1}}]}

\title{%
Evolutionary Feature Selection for Spiking Neural Network Pattern
Classifiers 
}

\author{\authorblockN{Michael Valko}
\authorblockA{
Department of Applied Informatics, \\Comenius University,\\ 
Mlynsk\'{a} Dolina, 842 48 Bratislava, \\Slovakia\\
Email: valko@sturak.sk}
\and
\authorblockN{Nuno C. Marques}
\authorblockA{CENTRIA,Departmento de Informatica,\\
Faculdade de Ciencias e Tecnologia,\\
University Nova de Lisboa,\\
Quinta da Torre,
2829-516 Caparica, \\ Portugal\\
Email: nmm@di.fct.unl.pt}
\and
\authorblockN{Marco Castellani}
\authorblockA{CENTRIA,\\
Faculdade de Ciencias e Tecnologia,\\
University Nova de Lisboa,\\
Quinta da Torre,
2829-516 Caparica, \\ Portugal\\
Email: mcas@fct.unl.pt}
}

\maketitle

\begin{abstract}
This paper presents the application of the biologically realistic JASTAP neural network model to
classification tasks.  
The JASTAP neural network model is presented as an alternative to the basic multi--layer perceptron model. 
An evolutionary procedure previously applied to the simultaneous solution of feature selection and neural
network training on standard multi--layer perceptrons is extended with JASTAP model.
Preliminary results on IRIS standard data set give evidence that this extension allows the use of smaller
neural networks
that can handle noisier data without any degradation in classification accuracy.
\end{abstract}


\section{Introduction}

Classification of patterns is important to many data mining processes.
Among possible classifiers, artificial neural network classifiers have proven to be one of the most robust classification systems. 
Their capability to deal with noisy input patterns, to handle both noisy and continuous value data
did prove to be an essential tool for classification and prediction~\cite{mitchell97machine}.

This paper presents a research aiming to
replacing the basic perceptron model~\cite{rumelhart1986learning} to the neural realistic JASTAP model~\cite{janco94modeling}, in some problems were a better representation of recurence and context may be needed. 
JASTAP model implements biologically plausible neural networks in a well parameterized model that
can be statistically related\cite{valko2005evolving}. Since the usual simplifications for other spiking models~\cite{maass1999pulsed} are not done a priori, we can keep all the flexibility known in biological neurons. 
On the same time, since parameters can also be predefined in advance (giving rise to more simple models) JASTAP proves as a very powerful new way of representing both model knowledge and experience. This way, only needed features can be activated in the model. A first demonstrative study on the use of this new kind of artificial neural network for classification tasks is presented on IRIS dataset.

Standard backpropagation learning~\cite{rumelhart1986learning} on JASTAP neural networks (and on simpler spiking models in general) is still an open issue (mainly due to the different weights that need to be adjusted and to cyclic connections). So, a more generic and powerful learning method is used: we have incorporated in JASTAP the evolutionary learning model of FeaSANNT~\cite{FeaSANNT}. FeaSANNT is an evolutionary procedure for simultaneous solution of the two combinatory search problems of feature selection and parameter learning in artificial neural network classifier systems.
FeaSANNT was already successfully applied to feature selection on traditional
artificial neural network pattern classifiers~\cite{FeaSANNT}.
In this paper we extend previous work by applying the JASTAP model to the same
type of classification problems.

An additional motivation for this work relies on the need to encode background information and recurence in 
neural networks. Indeed, recent work presents the so called
neuro--symbolic networks~\cite{JAL}. Based on a logic representation, these systems
provide both simple ways to represent predicate logic programs as MLP neural
networks~\cite{Hitzler2004} as well as to represent neural networks as rule base systems~\cite{JAL}.
Unfortunately, providing simple and adjustable structures
for encoding previous knowledge or algorithms in a neural network is a difficult task (e.x. Siegelmann's 
{\it Neural Automata and Analog Computational Complexity} article in \cite{Arbib2003}). Due to the simple model
for each unit, basic multilayer perceptron models often need many units (most of times organized in more than
one layer). 
Also, the simple MLP (and the related popular backpropagation learning
algorithm~\cite{rumelhart1986learning}), can not handle cyclic connections.
The well known usage of MLPs as universal approximatiors can only be achieved by means of enlarging the
hidden layer neurons as needed (e.x. K\"urkov\'a's {\it Universal Approximators} article in ~\cite{Arbib2003}).
As a result, in problems where context or time
representation is needed, repetive iterations can not be used, because there are no recursive structures in MLP.
Spiking neurons in general and JASTAP model in particular may present a solution for this problem.

The remaining of this paper is organized as follows. Section \ref{JASTAPS} presents the JASTAP spiking neural model.
Then the learning procedure used is presented in section \ref{FeaSANNTS}. The changes made to adapt this procedure
and some advantages and properties specific for JASTAP networks are described in section \ref{FeaSTAPS}.
Experimental results are presented in section \ref{sec:results}. Finally, the contributions of this work are
presented in section \ref{ConclusionsS}.

\section{\label{JASTAPS}JASTAP}%
\subsection{JASTAP As a Spiking Neuron Model}
JASTAP\footnote{pronounced as Yastap} model aims to simulate some biologically realistic 
functions of neural networks~\cite{janco94modeling}. JASTAP belongs to the family of
spiking models. Formal spiking neuron models 
in general work with temporal coding with mostly biologically relevant 
action potentials. Spiking models can also perform  
complex non--linear classification~\cite{bohte2002error}. This can be achieved with less 
units (spiking neurons) than with classical rate--coded 
networks (please see section ~\ref{subsec:results}).

In this section we briefly present JASTAP
model~\cite{janco94modeling} and the parameters used for easier adaptation of
this model to classification tasks. More details on the JASTAP model can be found
in~\cite{janco94modeling}.

\subsection{Model Description}

A JASTAP model is an artificial NN,
which consists of JASTAP neurons 
as the basic elements.
A neuron is described with:

\begin{itemize}
  \item \valkoitem{set of synapses}
    As it is usual in any neural network model, an JASTAP neuron is
    interconnected with its environment by one or
    more synaptic inputs and a single output (axon). The output can be connected
    with one or more neuron synapses in the network.

    For each synapse we consider:

    \begin{itemize}
      \item \valkoitem{input}
	--- can be internal (connected with axon of other neuron)
	or external (from the outer environment).

      \item \valkoitem{shape of postsynaptic potential (PSP) prototype} ---
	Biological neuron waveform evoked by a spike arriving 
	at a synapse is described in JASTAP model by

	\begin{equation}
	  \label{eq:psp}
	  \PSP(t) = k\cdot{\left(1-\e^{-\frac{t}{t_1}}\right)} ^2\cdot \e^{-\frac{2t}{t_2}}
	\end{equation}

	The waveform inter alia (controlled by parameters $t_1$ and $t_2$)
	emulates whether the synapse is located on a soma or on a 
	dendritic tree. $t_1$ and $t_2$ can vary from synapse to synapse. 
	They determine the potential decay.
	For example, \cite{redman83time} mentions $t_1 = 0.3\ms$ and $t_2=2.7\ms$.  As the neuron carries out the time--and--space summation of the input potentials and due to time discretisation during simulation, a more moderate decay
	can cause more sophisticated information processing. Therefore, we used $t_1$ up to $5\ms$ and $t_2$ up to $15\ms$ (please see figure~\ref{fig:psp}).

	\begin{figure*}
	  \begin{center}
	    \input{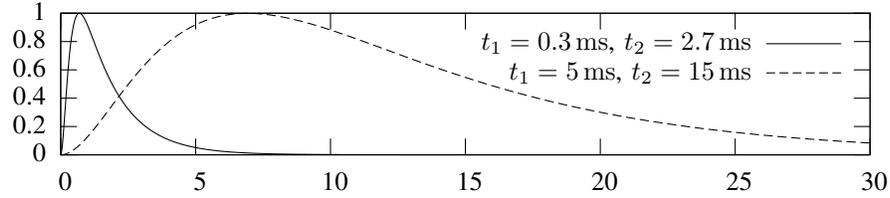}
	    \caption{Postsynaptic potential (PSP) with different waveform inter alia.}
	    \label{fig:psp}
	  \end{center}
	\end{figure*}

      \item \valkoitem{latency} ---  the time delay of the synaptic transmission
	and the axonal conduction. 
	In other words, synapse
	is silent for the latency time before it starts to translate the input impulse 
	into an action potential.	

      \item \valkoitem{synaptic weight (SW)} --- a value from 
	$\interval{-1}{1}$ that represents the strength of the synaptic input.
	An excitatory $\PSP$ is distint
	from a synapse with a positive SW
	and an inhibitory $\PSP$ from a synapse with a negative SW. 

      \item \valkoitem{plastic changes} --- Depending of the synapse type,
	SW can be influenced by some mechanisms, for instance Hebbian
	learning or 
	heterosynaptic presynaptic mechanism.
	In future this can be used to enable online learning. 
	In this work this feature of the model is not used,
	because the learning during information processing phase 
	is not our goal.

    \end{itemize}
  \item \valkoitem{instantaneous membrane potential (MP)} ---
    a quantity within the $\interval{-1}{1}$ range, determined
    as the sum of $\PSP$s limited by the non--linear function (figure~\ref{fig:atan})
    \begin{equation}
      \MP(t)=\frac{2}{\pi}\cdot\atan\left(\sum\nolimits_{\romantext{synapses}} \PSP(t)\right)
    \end{equation}

    \begin{figure*}[htbp]
      \begin{center}
	\input{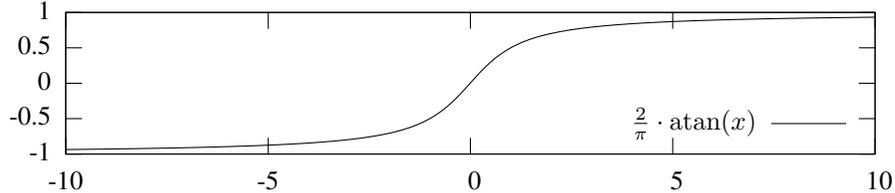}
	\caption{Limiting non--linear function}
	\label{fig:atan}
      \end{center}
    \end{figure*}

  \item \valkoitem{threshold} --- $\theta$, a value from $\interval{0}{1}$ ---  
    that determines the limit for firing.

  \item \valkoitem{spike frequency} --- the spike frequency 
    is restricted by the absolute refractory period. This is managed
    by setting minimum $I_{\min}$ and maximum $I_{\max}$ inter--spike interval for 
    the firing pattern.
    The standard value we use is  $I_{\min}=1\ms$ for the lowest and 
    $I_{\max}=10\ms$ for the highest value. 
    The actual inter--spike interval $I_a$ is determined as:
    \begin{equation}
      I_a = I_{\max} -(I_{\max} - I_{\min}) \cdot \frac2{\pi} \cdot  \atan\left(\frac{\MP-\theta}{1-\MP}\right)
    \end{equation}

    The spike frequency condition does not allow the neuron to fire sooner, 
    even if the MP exceeds the threshold.
\end{itemize}
\subsection{Tasks Where JASTAP Has Already Succeeded}
JASTAP was primarily designed to model real information 
flow in human brain. Until now, most experiments with JASTAP intend to prove this behavior.
On one hand, \cite{pavlasek2001temporal} shows that the JASTAP networks
can handle (even noised) information coded in temporal patterns.
On the other hand, \cite{pavlasek2003rate} shows its ability 
to recognize and distinguish different spike rates.
This is particularly interesting since the standard MLP weights
can be seen as the mean spiking rate concept. 

JASTAP  model also presents several capabilities that are related with the Gamma
distribution \cite{valko2005evolving}.
Gamma distribution 
was chosen for experiments because of  biological evidence of its
plausibility~\cite{koch98biophysics}. The results shown that the JASTAP models
are able to make decisions about features (mean rate, coefficient of variation)
of Gamma distribution even with very few neuron units. 
Moreover~\cite{valko2005evolving} explored evolved networks and described
at low--level how the actual decisions are made. 

\section{\label{FeaSANNTS}FeaSANNT}
\subsection{What Is FeaSANNT?}

FeaSANNT is an evolutionary procedure for simultaneous solution of the two
combinatorial search problems of feature selection and parameter learning for
artificial neural network classifier systems~\cite{FeaSANNT}.

One of the major problems in the use of ANNs is to train the frequently
large set of parameters
(usually, the connection weights).
Most of ANN weight training procedures are based on gradient descent of
the error surface, so they are prone to sub--optimal convergence to local
minima. 
Global search techniques such as evolutionary
algorithms (EAs) are known to produce more robust
results when pursuing multi--objective
optimization in large, noisy, multimodal and deceptive search spaces,
such as this one.

\subsection{Feature Selection in an Embedded Approach}

Feature selection can be regarded as a search
problem in the discrete space of the subsets of data attributes. 
Unfortunately, due to the often large set of attributes and their
interactions, selecting the optimal feature vector is usually difficult
and time consuming. Once again, the search space is noisy, complex,
nondifferentiable, multimodal and deceptive and the results are strongly
related 
to the type of classifier system used. However, as pattern classification
is based on the information carried by the feature vector,
feature selection has a crucial impact on the classifier accuracy, 
learning capabilities and size. 

FeaSANNT implements an embedded approach
in an evolutionary feature selection paradigm.
The search is guided by using GAs for learning how to
mask inputs. 
This paper reports on the extension of FeaSANNT model to handle
JASTAP Spiking Neural Networks. FeaSANNT's embedded approach was found as
particularly useful when a genetic algorithm was used 
for evolution as well as for learning. 
This way we were able to use the global nature of
the evolutionary search to avoid
being trapped by sub--optimal peaks of performance while learning
both the best set of features, neuron threshold, synapse weights and
synapse latencies. 
 
\subsection{What has been already done?}

Previous experiments with FeaSANNT ~\cite{FeaSANNT}, shown that the
simultaneous evolution of the input vector and the ANN weights
allows significant 
saving of computational resources. FeaSANNT algorithm was
applied on the popular multi--layer perceptron classifier.
Also several experiments were preformed on
six real--world numerical data sets that gave accurate and robust learning results. 
Significant reduction of the initial set of input features was achieved 
in most of the benchmark problems considered.
Examination of the evolution curves revealed selection of the optimal
feature vector takes place at an initial stage of the search. 
 FeaSANNT seemed also to compare well with other
classification algorithms in the literature.
However, the proposed algorithm entails lower computational costs due to
the embedded feature selection strategy.

\section{\label{FeaSTAPS}Using JASTAP NN model for Pattern Classification}
\subsection{Expected Benefits}
Expected benefits of spiking models, like JASTAP for pattern recognition (together with feature selection), include:

\paragraph{Smaller network structures} 
As stated previously, when compared to MLP, JASTAP neurons should require a
smaller number of units.
There is already some evidence that the decrease in number of neuron units can be seen even in the simple
XOR problem~\cite{bohte2002error} when using Spike Response Model~\cite{gerstner1995time}.
\paragraph{Noise filtering} In~\cite{pavlasek2001temporal} authors point out
that JASTAP is able to extract information with \emph{background spiking noise}.
In section~\ref{sec:results} it will shown how much noise JASTAP model can bear when applied to the IRIS dataset. 
We will be especially interested in processing data with 
Gamma noise, as this kind of noise that is biologically relevant~\cite{koch98biophysics}.
\paragraph{Results amenable to analysis} A smaller number of units and 
detailed processing (temporal coding instead of rate coding) give us a chance
to take an insight to internal operations at the level of synapses.
Simply speaking, we will try to 
\emph{decode} what is the evolved network actually doing. 
\subsection{Encoding Variables into Spike--Time Patterns}
\label{subsec:encoding}
Encoding processed data into time patterns is still an open issue even for Neurophysiology. 
For example, it is not clear if the temporal~\cite{singer1999time} or rate coding
representations play a crucial role in information processing~\cite{sejnowski1999neural}.

We found a first mention to this issue in \cite{bohte2002error}.
Authors used \emph{receptive fields} to transform real--world data into 
patterns. Although this approach can be neurologically plausible, for our (FeaSANNT) purposes
it was decided to use a \emph{repetitive} encoding
when the same inter--spike interval representing the data value is repeated as an 
input to input neuron. When the mask element is applied to a particular input, that neuron 
just receives an empty input train. 

\begin{figure*}
   \begin{center}
      \includegraphics[clip,viewport=148 650 528 708]{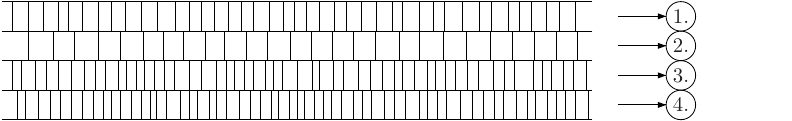}
      \caption{Example of encoding data to temporal code: 
      first iris--setosa training example (i.e. [5.1,~3.5,~1.4,~0.2]
      scaled to [7.2, 11.3, 5.7, 5.4]).
      Each row represents an input to 1 of 4 input neurons.
      Data are scaled to $5$--$15\ms$ and repeated over time period of 300~$\ms$.      
      Every interspike--interval is noised with $\pm 1 \ms$ of Gamma noise. 
      }
      \label{fig:encoding}
    \end{center}
    \end{figure*}

Every input is associated with one input neuron as an external synapse. 
Spike potentials for input neurons are precalculated from input data
in such a way that the inter--spike lengths are linearly dependent on
the data value (in 
the range of $5$--$15\ms$).
If no noise is taken into account interspike intervals corresponding
to one input train are the same. When we test the noise handling, each interspike
interval receives a value chosen from the Gamma distribution.
This kind of noise is considered as more biologically relevant (rather than applying noise to
a particular value) because it indicates cumulative noise during information
processing over several neurons (fig.~\ref{fig:encoding}).

Decoding to output can be simple. The following method is used:
Some neurons in the network are designated  as the \emph{output} neurons. 
When any of the output neurons fire, they are taken as a \emph{hot--spot}
network decision. The meaning of the decision is task--dependent.
Thereafter the simulation in progress can be stopped, because the subsequent
activity has no impact on the final decision.

\subsection{Designing Fitness Function for JASTAP Classification}
\label{subsec:fit}
During the tests preformed the selection of fitness function was a crucial step 
when evolving (recurrent) JASTAP networks. The fitness function was crucial not only for the number
of iterations for evolution (i.e. the time the network took to learn), but also in effectivity of
learning.

In the first place, the simple successful ratio based fitness functions
gave very poor results. A problem with these functions is that they cannot distinguish 
between networks with no response from those which are able to fire 
(even if not correctly). Furthermore local minima are common in these cases. 
This is the consequence of the emergence of the individuals responding to all 
inputs in the same way. 
To avoid such problems it was created a fitness function that favors the following criteria:
\begin{itemize}
  \item \emph{early} or non--silent \emph{responses} on input pattern 
can be classified as incorrect. 
      \item \emph{hetereogeneousness}: A network responding (correctly) 
	to many patterns from various classes is better that than a network
	that can correctly respond to only one. (With same overall ratio).
      \item \emph{selectivity}: we should evaluate with higher fitness networks responding mostly correctly 
	to one class, even if responding randomly (or not responding) to the other classes. Indeed these networks exhibit
        selectivity to some class are better that networks which respond to all patterns in the same 
	way (this is often case, when a population is initialized with random weights).
      \item For datasets with many classes, individuals responding in several classes should be evaluated with
	higher fitness. 
    \end{itemize}
    A linear combination of the several (sub--)fitness criteria stated above was used for the final 
  fitness function. First, by favoring small differences in correct responses:
    $$c_1/\Bigl(1+\varepsilon-\!\!\!\!\!\!\sum_{i\in\mathrm{classes}} \!\!\!\!\!\! \left(
    \mathrm{correct\mbox{--}ratio}[i]\right) \Bigr) $$
    Second, heterogeneousness and selectivity is mantained by emphasizing the
	minimal fitness, 
    i. e. the fitness based on minimal (with respect to classes) correct responses:
    $$c_2/\Bigl(1+\varepsilon-\min_{i\in\mathrm{classes}} \!\!\!(\mathrm{correct\mbox{--}ratio}[i])\Bigr)$$
    Finally, to handle and favor ability to respond in multiclassed datasets, the function gives little 
    credit when the network is able to respond correctly in any pair of classes.
    $$c_3\cdot\!\!\!\!\!\! \sum_{i \ne j \in \mathrm{classes}} \!\!\!\!\!\!
    \min (\mathrm{correct\mbox{--}ratio}[i],\mathrm{correct\mbox{--}ratio}[j])$$
    Experiments were carried out with the values $[c_1,c_2,c_3] = [1, 30, 7]$.
    %

\section{Results and Comparisons}
    \label{sec:results}
    \subsection{IRIS Data Set}
In order to evaluate results the JASTAP in FeaSANNT framework was tested on IRIS 
	UCI ML~\cite{Hettich+Blake+Merz:1998} well-known benchmark problem. 
 Inputs were converted into 
    temporal code as described in section \ref{subsec:encoding}.

The Iris data set was firstly presented 
in a statistical study for determining the classification of three sets of flowers. The classification set has three distinct classes of plants and four numeric features (corresponding to the petal and sepal width and length). A sample of $150$ examples were collected ($50$ for each class). Only one class is linearly separable from the remaining ones, because some data points on the other two classes are intersecting each other (however most of data points are fairly distinct). From statistical evidence only the measures of petal width and length are enough for classification.


In FeaSANNT algorithm only two features were selected with an average accuracy of $94.7$. By using the four features with an MLP with backpropagation learning $96.2\%$ of accuracy were achieved.


    \begin{table*}
      \begin{center}
	\begin{tabular*}{0.8\textwidth}{@{\extracolsep{\fill}}|c|c|c|c|c|c|}
	  \hline
	  \textbf{name} & \textbf{source} &  \textbf{size} &  \textbf{features} &  \textbf{classes} &  \textbf{training set} \\
	  \hline
	  Iris & UCI ML & 150 & 4 & 3 & 80~\% -- random \\
	  \hline
	\end{tabular*}
	\caption{Data sets}
	\label{tbl:datasets}
      \end{center}
    \end{table*}


    \subsection{Set--Up} 
    \subsubsection{Chromosomes}
    Structure is not evolved so the individual consists of chromosome defining 
    values subject to evolution and the binary mask for feature selection.
    Gray binary coding was used for the chromosome network values. The evolution 
    of neuron thresholds and synapse weights and latencies is crucial 
    for network function. Following results of~\cite{valko2005evolving}, JASTAP is not
    evolved for $\PSP$'s $t_1$ and $t_2$ parameters $I_{\min}$ and $I_{\max}$ defining bounding 
    firing periods. Table~\ref{tbl:bound} displays used JASTAP setup.
    \begin{table}
      \begin{center}
	\begin{tabular}{|c|c|c|}
	  \hline
	  \textbf{feature} & \textbf{min value} &  \textbf{max value} \\
	  \hline
	  synapse weight  & $-1$ & $1$ \\
	  \hline
	  synapse latency & $0\ms$ & $40\ms$ \\
	  \hline
	  synapse waveform $t_1$ value & \multicolumn{2}{|c|}{$5\ms$} \\
	  \hline
	  synapse waveform $t_2$ value & \multicolumn{2}{|c|}{$15\ms$} \\
	  \hline
	  neuron threshold & $0$ & $1$ \\
	  \hline
	  minimal neuron firing period & \multicolumn{2}{|c|}{$1\ms$} \\
	  \hline
	  maximal neuron firing period & \multicolumn{2}{|c|}{$10\ms$} \\
	  \hline
  	  maximal network response time & \multicolumn{2}{|c|}{$300\ms$} \\
	  \hline

	\end{tabular}
	\caption{Parameter boundaries used for evolution}
	\label{tbl:bound}
      \end{center}
    \end{table}

    \subsubsection{Evolutionary Algorithm}
    Selection is made in an elitism--like manner: Recombined individuals are evaluated with the same 
    examples as their parents and then the best part (a half) of all chromosomes (parents and offspring together)
    is taken as the next population.

    \begin{table}
      \begin{center}
	\begin{tabular}{|c|c|}
	  \hline
	  \textbf{feature} & \textbf{value}  \\
	  \hline
	  trials & $1$ \\
	  \hline
	  generations & $100$--$500$ \\
	  \hline
	  population size& $100$ \\
	  \hline
	  values crossover rate& $0.7$\\
	  \hline
	  values mutation rate& $0.05$ \\
	  \hline
	  mask mutation rate& $0.05$ \\
	  \hline
	\end{tabular}
	\caption{Parameter settings of evolutionary algorithm}
	\label{tbl:evo}
      \end{center}
    \end{table}
    \subsection{Results}
    \label{subsec:results}
    \subsubsection{Iris data set} 

\begin{figure} 
\begin{center}
\includegraphics[viewport=159 585 248 700]{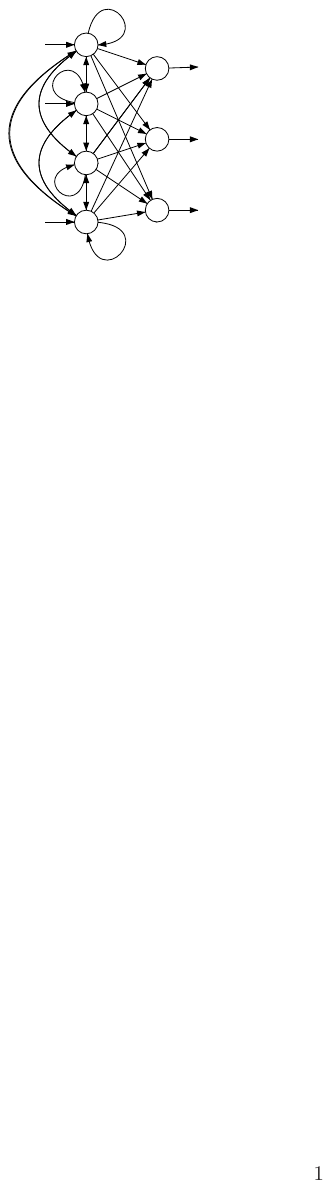}
\end{center}
\caption{ Iris network}
\end{figure}

This experiment on IRIS data set aimed at showing the potential of JASTAP due 
to the smaller number of neuron units needed for good classification. Therefore the hidden layer was totally dropped 
    leaving only 7 neurons in structure (4 for inputs and 3 for outputs).
    In addition we interconnected neurons in input layer to enable mutual information exchange and recurrence.
    The results are shown in table~\ref{tbl:iris}. 
    BPrule stands for classical BP algorithm~\cite{rumelhart1986learning}.
    ANNT and FeaSANNT refer to the EA algorithm either disabled or enabled 
    feature selection module~\cite{FeaSANNT}. Present results are in FeaSTAP\footnote{FeaSTAP means JASTAP in FeaSANNT} column.
    \begin{table*}
      \begin{center}
	\begin{tabular}{|c|c|c|c|c|}
	  \hline
	  \textbf{iris} & \hbox to 6em {\textbf{BPrule}}  &  \hbox to 6em {\textbf{ANNT}} &  \hbox
	  to 6em {\textbf{FeaSANNT}} &  \hbox to 6em {\textbf{FeaSTAP}} \\
	  \hline
	  hidden layer size & 2 & 2 & 2 & no \\
	  inputs & 4 & 4 & 1.5 & 3\\
	  accuracy & 96.2~\% & 95.3~\% & 94.7~\% & $100~\%$\\
	  iterations & 700 & 300 & 400 & 300\\
	  \hline
	\end{tabular}
	\caption{Iris data set}
	\label{tbl:iris}
      \end{center}
    \end{table*}
	  \subsubsection{Handling the noise}
	  JASTAP model was designed to be as detailed as it is needed to 
	  simulate biorealistic functions in reasonable way. With this in mind,
	  we have conducted experiments to test the noise handling
	  during classification (please see subsection ~\ref{subsec:encoding}). 
	  To keep on being biologically inspired we have used
	  the noise generated from Gamma distribution. 
	  The noise value in cortical spike times
	  is about $1$--$2~\ms$ \cite{maass1999pulsed}
	  what is about $10~\%$ due to our encoding. 
	  Such a noise can be generated from 
	  $\mathscr{G}(\alpha=25,\beta=0.8)$
	  Smaller amounts of noise (at the $1~\%$ level) were also tested.

	  Gamma distribution generators GS$^\ast$ and  GKM1
	  from~\cite{fishman96monte} were used. These generators use a combination of 
	  the acceptance--rejection, composition and inverse transform methods.

	  \begin{table}
	    \begin{center}
	      \begin{tabular}{|c|c|c|c|}
		\hline
		\textbf{Noised Iris} &\hbox to 6em { no\ noise }  &\hbox to 6em { $1\% \approx  0.1
		\ms$ } & \hbox to 6em { $10~\% \approx  1 \ms$ } \\
		\hline
		hidden layer size & no & no & no \\
		inputs &  3  & 3  & 3 \\
		accuracy & 100.0~\% & 100.0~\% & 100.0~\% \\
		iterations & 200 &  200  & 300\\
		\hline
	      \end{tabular}
	      \caption{Noised Iris data set}
	      \label{tbl:niris}
	    \end{center}
	  \end{table}

This results show that noise levels until the $10~\%$ level don't reduce the
classification quality. However the problem becomes more dificult to learn on
the $10~\%$ noise level. By increasing the noise level beyond this limits
classifier accuracy starts being compromised. 
  It should be noted that noise was applied to each inter--spike interval independently (even in the same input train).


	  \section{Conclusions}
\label{ConclusionsS}
This paper main goal is to show that the basic perceptron model can be replaced by the biologically realistic JASTAP
neural network model for classification tasks. In order to do so, an evolutionary procedure for simultaneous
solution of feature selection and for JASTAP neural network parameter adjusting was used.

Preliminary experimental results show that not only JASTAP seems to be able to deal with the same
learning problems with smaller neural networks, but also JASTAP unique features enable the
insertion of artificially generated noise into the learning set without degrading learning performance. 
Indeed, results on the Iris 
standard data set
\cite{Hettich+Blake+Merz:1998} use smaller neural networks without compromising accuracy. Also, noise was 
artificial inserted into training data without any degradation in classification accuracy.
However, it was noticed that the feature selection (reduction of mask size) was not 
so significant (it has always 3 or 4 mask elements) as in original FeaSANNT work. 
We assign it to the fact, that the JASTAP network use the spikes potentials
even from irrelevant inputs just to overcome the threshold.
This is also probably the reason why $100\%$ precision is achieved on the IRIS data-set.
Since this data set is known not to be fully separable, this high value is probably due
to a statistical abnormality either in sample data or in the evaluation procedure.
Anyway care should be taken on future experiments to avoid opportunistic overlearning
strategies.

Although not discussed in the paper, the behavior of the evolved networks for IRIS
classification was also studied. Indeed, it was observed that the networks do the classification
by emphasizing differences in the features correlating with the classes {\it petal length} and {\it width}.

A major problem with JASTAP learning model was the computational time it took to learn. This paper is
presenting a learning model for a difficult problem: there are several parameters to tune, time is directly
simulated and we are performing feature selection. In future work 
we hope to overcome some of these problems by improving the time simulation and by restricting more the
parameters that should be learned. Also running FeaSANNT procedure in parallel should also help to
reduce learning times.

We hope to use JASTAP neural networks to represent logic models as neural networks. 
In such a neural network, the architecture and some set of weights and parameters will be fixed for
encoding background information. For that the relations of the JASTAP model with Gamma
distribution \cite{valko2005evolving} should be studied. The small number of neurons needed for data
processing in JASTAP models could
help to make the network more modular.

Several other features of JASTAP model can also be used with advantage for classification tasks.
For example the ability for plastic changes (already studied in JASTAP) can be very interesting for
online adaptation of the neural network classifier to small changes in the learned model. 
Because the JASTAP model
works with temporal code, we argue that it can more easily discover statistical regularities over
longer time in input pattern trains and encode them on the output. This could be useful in
classification problems where context is an issue (e.g. \cite{MarquesLopes2001}, \cite{CastellaniMarques2005}).




%

\end{document}